# Using Potential Influence Diagrams for Probabilistic Inference and Decision Making

**Ross D. Shachter**
Department of Engineering-Economic Systems
Stanford University
Stanford, CA 94305-4025
shachter@camis.stanford.edu

**Pierre Ndilikilikesha**
Fuqua School of Business
Duke University
Durham, NC 27706
ndili@dukefsb.bitnet

## Abstract

The potential influence diagram is a generalization of the standard "conditional" influence diagram, a directed network representation for probabilistic inference and decision analysis [Ndilikilikesha, 1991]. It allows efficient inference calculations corresponding exactly to those on undirected graphs. In this paper, we explore the relationship between potential and conditional influence diagrams and provide insight into the properties of the potential influence diagram. In particular, we show how to convert a potential influence diagram into a conditional influence diagram, and how to view the potential influence diagram operations in terms of the conditional influence diagram.

## 1 INTRODUCTION

The potential influence diagram (PID) is a generalization of the standard influence diagram, a directed network representation for probabilistic inference and decision analysis [Ndilikilikesha, 1991]. Instead of factoring a joint distribution of the variables into conditional distributions, the PID only requires that the joint distribution be factored into nonnegative components. This small change allows more efficient computation and updating from evidence observation, but it also completely changes the semantics, independence, and fundamental operations of the influence diagram. To emphasize the contrast between the PID and the standard influence diagram, we shall refer to the standard diagram as a conditional influence diagram (CID). In this paper, we explore the relationships between the CID and the PID.

A major motivation for using the PID representation is efficient computation. The algorithms for probabilistic inference and decision analysis on the CID require normalizing divisions in order to maintain the conditional structure [Shachter, 1986; Shachter, 1988; Shachter, 1989]. More efficient algorithms maintain an undirected graphical structure in which such normalization operations are unnecessary [Jensen et al., 1990a; Jensen et al., 1990b; Lauritzen and Spiegelhalter, 1988; Shachter and Peot, 1992; Shafer and Shenoy, 1990; Shenoy, 1991; Shenoy, 1992]. Although the computational complexity of both approaches is of the same order, the latter methods obtain a constant improvement by avoiding the divisions and by maintaining fewer tables.

The main advantage of the CID over the undirected graphical methods has been the success of the directed graph as a representation for communicating a model. It has been proven effective for eliciting and presenting models among quantitative and nonquantitative parties. The PID allows the same directed graph to be used for the more efficient computations. Starting with a CID, a special case of the PID, various operations might transform the model to a PID that is not a CID. In this paper, we develop an algorithm to convert back to the CID, if it is needed, and we explore the general relationships between the two types of models. We do this by first developing the essential primitive operations on the PID that differ from the CID. This helps explain the nature of the more complex operations on the PID.

In Section 2 we review the basic properties and operations of CID's, and in Section 3 we introduce the corresponding properties of PID's. In Section 4, we combine these basic operations into more complex transformations and algorithms. Finally, in Section 5 we present some conclusions and suggestions for future research.

## 2 CONDITIONAL INFLUENCE DIAGRAMS

A conditional influence diagram is a network built on a directed acyclic graph. The nodes in the diagram correspond to uncertain quantities, some of which can be observed, decisions which are to be made, and the criterion for making those decisions. The arcs indicate the conditioning relationships among those quantities and the information available at the time the decisions must be made.

A **conditional influence diagram (CID)** is a network structure built on a directed acyclic graph [Howard and Matheson, 1984; Olmsted, 1983; Shachter, 1986]. Each node j in the set $N = \{1, \ldots, n\}$ corresponds to a variable $X_j$, and the nodes are partitioned into sets C, D, and V, corresponding to chance nodes (random variables) drawn as ovals, decision nodes (variables under the decision maker's control) drawn as rectangles, and the value node (criterion to be maximized in expectation when making decisions) drawn as a rounded rectangle. Each variable $X_j$ has a set of **possibilities**. Random variable (and value) $X_j$ has a conditional probability distribution over those possibilities; the conditioning variables have indices in the set of **parents or conditional predecessors** C(j), and are indicated in the graph by arcs into node j from the nodes in C(j). Each random variable $X_j$ is initially unobserved, but at some time its value $x_j$ might become known. At that point it becomes an **evidence variable**, its index is included in the set of evidence variables E, and this is represented in the diagram by drawing its oval with shading. The parents of a decision node $X_j$ have a completely different meaning; they are **informational predecessors** I(j), indicating which variables will be observed before the decision choice must be made.

As a convention, lower case letters represent specific possibilities and single nodes while upper case letters represent variables and sets of nodes. If J is a set of nodes, then $X_J$ denotes the vector of variables indexed by J. For example, the conditioning variables for $X_j$ are denoted by $X_{C(j)}$.

In addition to the parents, we can define the **children** of a node, its **ancestors**, and its **descendants**. A list of nodes is said to be **ordered** if none of the ancestors of a node follows it in the list. Such a list exists if and only if there is no directed cycle among the nodes.

The conditional distributions in the influence diagram factorize a joint distribution,

$$\Pr\{X_C = x_C, X_V = x_V \mid X_D = x_D, X_E = x_E\}$$
$$= \Pi_{i \notin D} \Pr\{X_i = x_i \mid X_{C(i)} = x_{C(i)}\},$$

that is, the joint distribution of the unobserved variables, conditioned on the decisions and the observations, is obtained by multiplying the tables for all nondecision variables. As a result, whenever random variable $X_j$ has no observed descendants, its posterior distribution can be obtained completely from its ancestors (and itself); on the other hand, when it has observed descendants, some of the information to compute its posterior distribution might be contained in its descendants.

There are several desirable conditions which simplify management of a CID while not restricting the sensible models that can be represented. When these conditions are satisfied, the CID is said to be **regular** [Shachter, 1986]. These conditions are:

1) **no directed cycles**: this simplifies the factoring of the distribution and prevents us from learning anything about a decision we have yet to make. Formally, this latter condition means that until an optimal policy for decision $X_d$ is determined, $X_d$ must be independent of $X_{I(d)}$. This is enforced in the structure of the diagram by not allowing any descendant of a decision to be a parent of the decision.

2) **total ordering of decisions**: this avoids ambiguities (we also assume the no forgetting condition, that at any decision we observe all information available at the time of earlier decisions, including our choice for the earlier decision);

3) **the presence of a value node if there are decisions**: this provides a criterion; and

4) **no children of a value node**: this simplifies the solution procedure.

We will assume for the rest of this paper that we are dealing with a regular influence diagram.

There are four primitive operations on the CID which are of interest in this paper: arc reversal, barren node removal, optimal policy selection, and evidence instantiation. (The other primitives involving deterministic nodes and expected value are computationally significant but distracting from the issues we are addressing here.) These four operations are illustrated in Figure 1. The conditional distributions, where significant, are labeled "(C)."

**Arc reversal**, shown in parts a) and b) of Figure 1, is the CID representation for Bayes' Theorem. Given a model in which $X_h$ is conditioned by $X_i$ and other variables, we transform our model into one in which $X_i$ is conditioned by $X_h$; in the process, each inherits the other's parents in the model,

$$\Pr\{X_h \mid X_J, X_K, X_L\}$$
$$\leftarrow \Sigma_{x_i} \Pr\{X_i \mid X_J, X_K\} \Pr\{X_h \mid X_i, X_K, X_L\}$$

and

$$\Pr\{X_i \mid X_h, X_J, X_K, X_L\}$$
$$\leftarrow \frac{\Pr\{X_h \mid X_i, X_K, X_L\}}{\Pr\{X_h \mid X_J, X_K, X_L\}}.$$

Arc reversal can be performed only when the (i, h) arc is the only path from i to h; otherwise, a directed cycle would be created. It is the normalization divisions in the arc reversal operation that we can avoid through the generalization to a potential influence diagram.

**Barren node removal**, shown in parts c) and d) of Figure 1, is the elimination of a nuisance variable, one that we neither care about nor observe. If an unobserved, non-value variable $X_i$ has no children, then we can eliminate it from the model. If it is a random variable (with a conditional distribution), then it contributes to no other posterior distributions; if it is a decision variable, then any choice would be equally desirable.

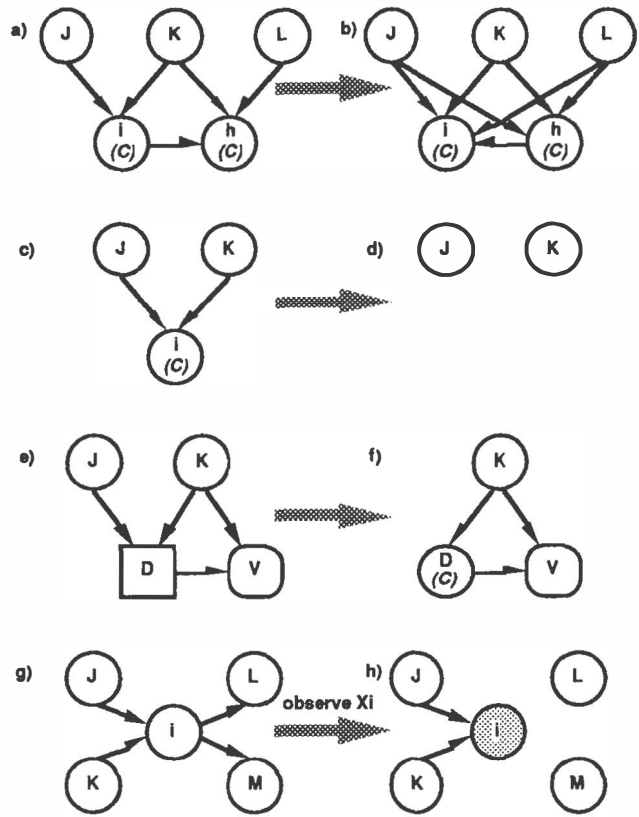

Figure 1. Primitive Operations for the Conditional Influence Diagram

Optimal policy selection, shown in parts e) and f) of Figure 1, is the substitution of a policy chance variable in place of a decision variable, once all of the uncertainties relevant to but unobserved before the decision have been removed from the problem. At that point, all of the parents of the value node are observed at the time of the decision,

$$C(V) \subseteq \{d\} \cup I(d),$$

and we can determine the optimal policy by

$$x_d^*(x_{I(d)}) = \max_{x_d} E\{ X_V \mid x_{C(V)} \}.$$

Evidence instantiation, shown in parts g) and h) of Figure 1, is the incorporation of an observation into the directed graph. If $X_i$ has been observed to be $x_i$, there is no longer any need to carry around the information about the other possibilities for $X_i$. As a result, all of the tables indexed by $X_i$ can be reduced; these tables are in the node i and its children. Therefore, after evidence instantiation the arcs from i to its children can be eliminated. If, on the other hand, the observation about $X_i$ is imperfect, this can be modeled by adding a new "dummy" child for i, observing that child exactly, and using evidence instantiation on that node.

## 3 POTENTIAL INFLUENCE DIAGRAMS

In this section, we relax the requirements for the conditional influence diagram slightly, to obtain the potential influence diagram. We show the two new primitives needed to exploit this generalization, and explore some of the properties of this new representation.

A potential influence diagram (PID) is a network structure similar to the CID, except that the tables for random variables and the value, now called **potentials**, $\text{Pot}\{X_i | X_{C(i)}\}$, are no longer necessarily conditional probability distributions [Ndilikilikesha, 1991]; all that is required is that they are nonnegative and that their product continues to satisfy

$$\Pr\{ X_C=x_C, X_V=x_V \mid X_D=x_D, X_E=x_E \}$$
$$= \Pi_{i \notin D} \text{Pot}\{ X_i=x_i \mid X_{C(i)}=x_{C(i)} \},$$

that is, the joint distribution of the unobserved variables, conditioned on the decisions and the observations, is obtained by multiplying the tables for all nondecision variables. Since this condition was satisfied in the CID, a CID is a special cased of a PID. In a potential influence diagram, some of the information to compute the posterior distribution for a random variable $X_j$ might be contained in its descendants, even when none of those descendants is observed. We say that a PID is **regular** if it satisfies the CID regularity conditions and there is some corresponding regular CID. We will hereafter assume that all PID's are regular without any loss of generality, since we can represent all sensible models using regular diagrams.

### 3.1 PRIMITIVE OPERATIONS

There are two primitive operations on the PID which allow us to exploit its special properties: potential reversal and conditionalization. These operations are shown in Figure 2. The potential distributions, where significant, are labeled "(P)" and empty distributions are labeled "(1)." Two of the primitive operations for the CID continue to work properly for the PID, optimal policy selection and evidence instantiation; there is nothing surprisingly about evidence instantiation, but there are some subtleties in the optimal policy selection that make it worth examining later in this section.

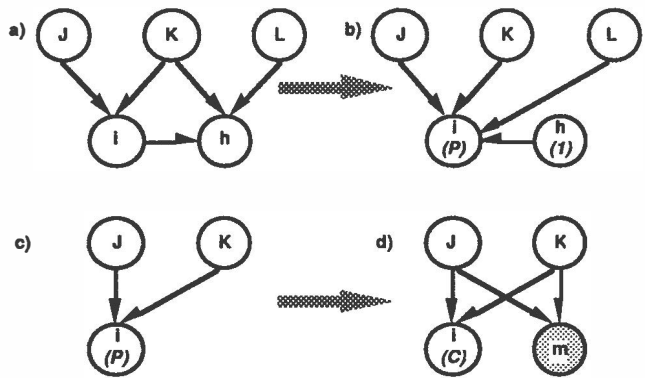

Figure 2. Primitive Operations for the Potential Influence Diagram

**Potential reversal**, shown in parts a) and b) of Figure 2, is the PID analog to arc reversal in the CID. Because all of the dimensions needed to store the product of the two potentials are available afterwards in node i, there is no longer a need to store any information in node j afterwards, and thus no need to send arcs into node j. The ability to store all of the information in node i is the key to the potential influence diagram. The new potentials are

$$\text{Pot}\{ X_i \mid X_h, X_J, X_K, X_L \}$$
$$\leftarrow \text{Pot}\{ X_i \mid X_J, X_K \} \text{Pot}\{ X_h \mid X_i, X_K, X_L \}$$

and

$$\text{Pot}\{ X_h \} \leftarrow 1 .$$

Potential reversal, like arc reversal, requires that the (i, h) arc is the only path from i to h; otherwise, a directed cycle would be created. The difference between potential reversal and arc reversal is that the product of the two potentials created is maintained (in node i) without normalizing divisions. Only one of the resulting tables is needed, so there are less numbers to maintain and arcs are only added to one of the nodes. The other has a scalar "1" and no incoming arcs at all!

**Conditionalization**, shown in parts c) and d) of Figure 2, is the means by which a potential distribution is transformed into a conditional distribution. Since barren node removal can be performed only on nodes with conditional distributions, this operation is required (implicitly) in order to eliminate unobserved variables from a model. The conditionalization operation normalizes a distribution to make it a valid conditional distribution, and creates a dummy observation node $X_m = x_m$ in which it stores the normalization constant. The calculations are

$$\text{Pot}\{ x_m \mid X_{C(i)} \} \leftarrow \Sigma_{x_i} \text{Pot}\{ X_i \mid X_{C(i)} \}$$

and

$$\text{Pr}\{ X_i \mid X_{C(i)} \} \leftarrow \frac{\text{Pot}\{ X_i \mid X_{C(i)} \}}{\text{Pot}\{ x_m \mid X_{C(i)} \}} .$$

As suggested by the conditionalization operation, there is a strong connection between the PID and observed evidence nodes in the CID. As an example, the three diagrams shown in Figure 3 are all "equivalent."

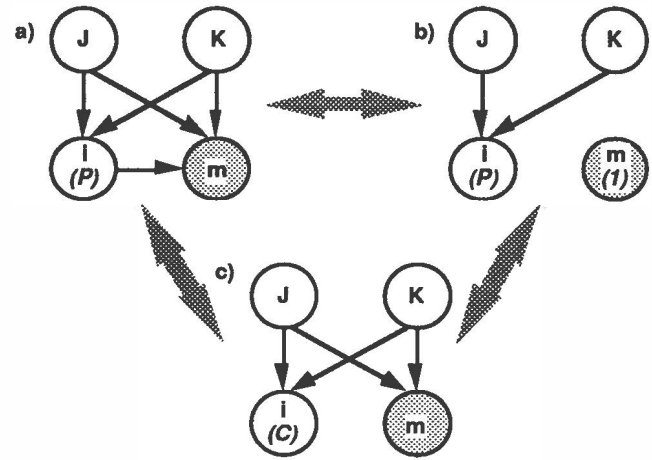

Figure 3. "Equivalent" Potential Influence Diagrams

**Theorem 1.**

The three diagrams shown in Figure 3 correspond to the same joint probability distributions and show the same conditional independence.

Proof:

We can get from part a) to part b) in Figure 3 by applying the potential reversal operation. (Note that because m is observed, it does not need to send an arc back to i.) We can get from b) to c) by applying the conditionalization operation. Finally, we can recognize that c) is just a special case of a), since we can always add an arc from i to m, and think of the conditional distribution at i as a potential distribution. #

### 3.2 INDEPENDENCE AND INFORMATION REQUIREMENTS.

As we have seen, the conditionalization operation is important as a graphical operation, as well as a numerical one. Another application of this operation is to verify the independence and information requirements properties of any inference or decision problem after adding an observation for each potential distribution [Geiger et al., 1989; Geiger et al., 1990; Shachter, 1988; Shachter, 1990]. These procedures can be made even more efficient; we get equivalent results if we simply add a single dummy observed node child to each node with a potential distribution, before checking for independence and information requirements.

## 3.3 OPTIMAL POLICY SELECTION

Finally, it is necessary to verify the optimal policy selection operation for the PID.

**Theorem 2.**

The optimal policy selection operation as presented for the CID works correctly for a regular PID.

Proof:

We give a graphical proof as shown in Figure 4. In part a) we satisfy the normal CID conditions that all conditioning variables for the value are observed at the time of the decision. We now use conditionalization to make the value node conditional. Normally, this requires adding an arc from D to m, creating a cycle, since m must be observed before the D decision. Recall, however, that in any regular CID corresponding to the regular PID (and in any sensible model), the optimal choice (before it is made) must be independent of all observations, so no arc is needed from D to m. We can now select the optimal policy, to obtain the diagram in part c), and then apply Theorem 1 to obtain the diagram in part d). Since the constant nonnegative potential stored in node m does not change the ordering of optimal policies, we could equivalently make the optimal choice directly on the value potential. #

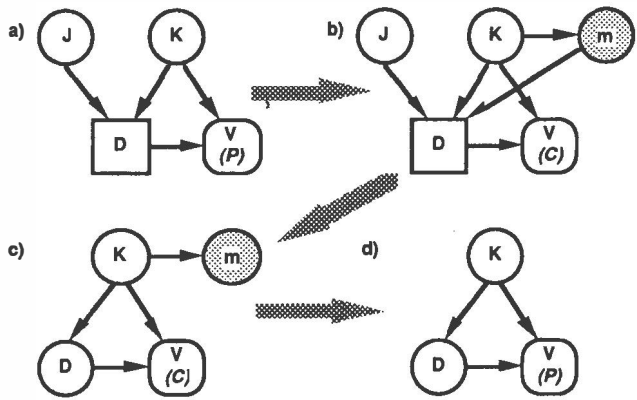

Figure 4. Derivation of Optimal Policy Selection in the Potential Influence Diagram

## 4 COMPOUND OPERATIONS AND ALGORITHMS

Having built the primitive operations for the PID, we can combine them to construct the compound operations and algorithms to solve inference and decision problems and to transform PID's into CID's.

### 4.1 EVIDENCE PROPAGATION

The first step in this process is the operation of **evidence propagation**, whereby a sequence of arc reversal operations is used to move all of the observed evidence nodes to the start of an ordered list of all nodes in the diagram [Chyu, 1990a; Chyu, 1990b; Shachter, 1989; Shachter et al., 1990]. This yields a diagram that represents the posterior joint distribution given the evidence. The algorithm visits each unobserved node in reverse order, according to some target order, and applies the following steps:

1. If there are multiple observed children for this node, combine them into a single observation child whose parents are the union of the parents of the original children; and

2. If there is an evidence child for this node, perform arc reversal from this node to the child. There is no need to create an arc back from the evidence afterwards.

Afterwards, the ancestors of each evidence node become decomposable, that is, if any two nodes share a common child then there is a directed arc between them; also, there is a unique ordered list of the ancestors of each evidence node after evidence propagation.

An example of evidence propagation is shown in Figure 5. The original graph, with five variables, is shown in part a). In part b) we have been given imperfect evidence about two of the variables, B and E, and now we can apply the algorithm with target order ( A B C D E ). Visiting E, reverse arc (E, G) to obtain the graph shown in part c). Now reverse the arc (D, G), and combine F and G into a single observation node with parents B and C, to obtain the graph shown in part d). If we continue, reversing the arcs (C, FG), (B, FG), and (A, FG), we end up with the graph shown in part e). Note that there is now a unique ordered list for the whole graph (all of the unobserved nodes were ancestors of G), and it is decomposable.

Alternatively, we could have started with target order ( C A B D E ). In this case, we should first reverse arc (A, C) so that the target order is ordered for the original graph as shown in part f). Now propagating the evidence F and G, we obtain the diagram shown in part g). Again, it is decomposable for the ancestors of G, but now those ancestors do not include A and so we obtain a simpler final graph. By choosing the "right" target order, we can obtain any desired decomposable graph [Chyu, 1990a; Chyu, 1990b; Shachter et al., 1990].

### 4.2 PID TO CID CONVERSION.

This same technique can be employed to convert a PID to a CID. Suppose that A, B, and C have conditional distributions, but D and E have potential distributions. We can conditionalize D and E to obtain the graph shown in part c), and apply evidence propagation to end up with either of the graphs shown in parts e) and g). A more efficient version would wait to conditionalize a node until it is visited in reverse target order.

normalizing constant before we remove i. The calculation is

$$\text{Pot}\{ x_m \} \leftarrow \Sigma_{x_i} \text{Pot}\{ X_i \}.$$

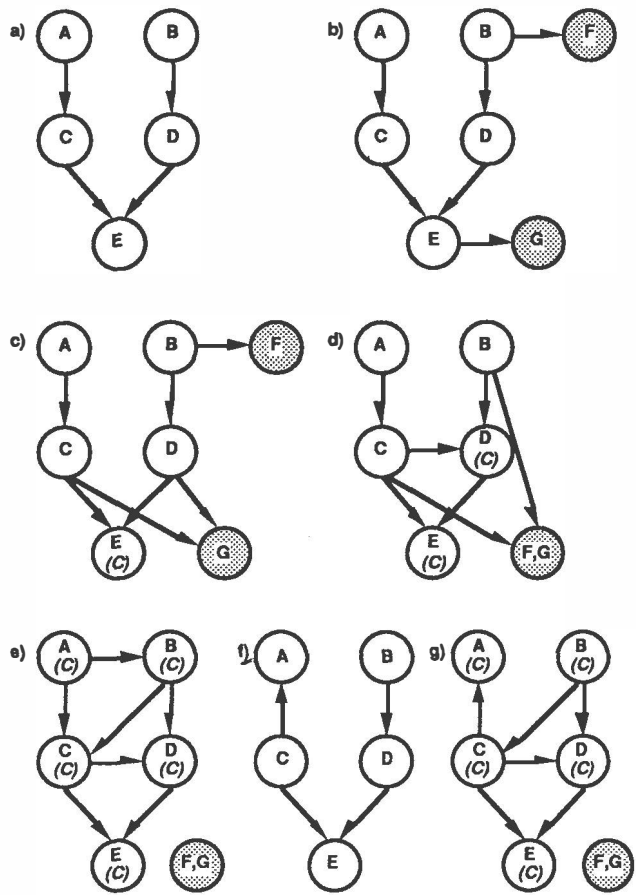

Figure 5. Evidence Propagation in a Conditional Influence Diagram

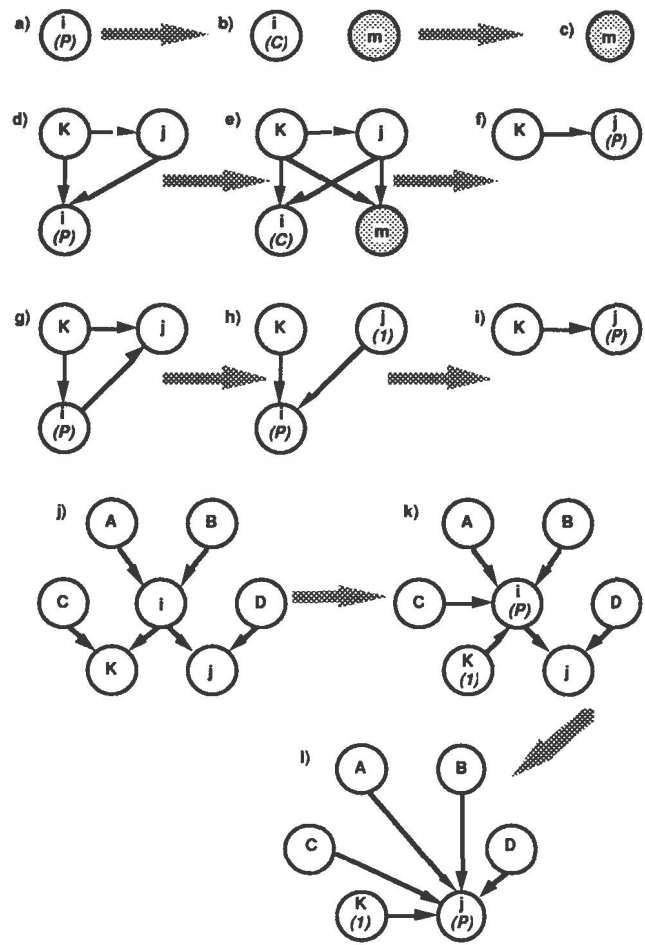

Figure 6. Probabilistic Reduction in a Potential Influence Diagram

The general operation of **probabilistic reduction** is the removal from the diagram of an unobserved random variable. In the CID, this is performed by reversing all of the outgoing arcs from a node (in the graph order of the children) until the node is barren and it can be removed. If it has a value child, that can always be last in the graph order, so it will be the last to be removed. Some added efficiency can be obtained in the last reversal and removal by recognizing that a new distribution does not need to be computed in the node that will be removed.

The probabilistic reduction is more complex in the PID, but there is greater opportunity for efficient computation [Ndilikilikesha, 1991]. The full story is shown in Figure 6 and with the four cases below for the unobserved node i that we want to reduce.

1. **no parents or children**: we take the diagram as shown in part a), conditionalize to obtain the diagram shown in part b), then apply the barren node removal to obtain the diagram in part c). The conditional distribution doesn't need to be built--we just need to keep track of the

2. **no children but at least one parent**: unlike the CID, the PID requires us to extract any information from node i about its parents before we can remove it. Distinguish one of the parents of i, j, such that there is no path from j to the rest of i's parents, K. (There might be a path from K to j.) Starting with the diagram shown in part d), we conditionalize to obtain the diagram in part e), so we can now perform barren node removal on i. Finally, we reverse the (j, m) arc as in Theorem 1, to obtain the diagram shown in part f). The calculation is

$$\text{Pot}\{ X_j | X_K \}$$
$$\leftarrow \Sigma_{x_i} \text{Pot}\{ X_i | X_j, X_K \} \text{Pot}\{ X_j | X_K \}.$$

3. **one child**: starting with the diagram in part g), use potential reversal to obtain the diagram in part h), and

then apply the no children case above to obtain the diagram in part i). The calculation is similar to the one above,

$$\text{Pot}\{ X_j \mid X_K \}$$
$$\leftarrow \Sigma_{x_i} \text{Pot}\{ X_i \mid X_K \} \text{Pot}\{ X_j \mid X_i, X_K \} .$$

4. **multiple children**: this is the most general case. First, pick one of i's children, j, which can appear after all of the others; if i is a parent of the value node, then the value node should be selected. Starting with the diagram as shown in part j), reverse the arc from i to each child (except j) to obtain the diagram in part k). Finally, apply the single child case above to obtain the diagram in part l). The calculations can be summarized as

$$\text{Pot}\{ X_j \mid X_{ABCDK} \}$$
$$\leftarrow \Sigma_{x_i} \text{Pot}\{X_i \mid X_{AB}\} \text{Pot}\{X_j \mid X_{iD}\} \text{Pot}\{X_K \mid X_{iC}\}$$

and

$$\text{Pot}\{ X_k \} \leftarrow 1 \text{ for all } k \in K .$$

### 4.3 PROBABILISTIC INFERENCE.

All the pieces are now in place to perform probabilistic inference in the PID as in the CID [Shachter, 1988], but with greater efficiency. Any observations of evidence should be instantiated and nuisance variables reduced; the resulting PID is the posterior diagram. If desired, it can be converted into a CID for communication to experts and decision makers. A message passing algorithm could also be developed for propagation of posterior marginals similar to [Jensen et al., 1990a; Jensen et al., 1990b; Shafer and Shenoy, 1990].

### 4.4 DECISION ANALYSIS.

In a similar fashion, all the pieces have been assembled to perform decision analysis as in the CID, and the same completeness proof applies [Shachter, 1986]. At each step in the algorithm, until there are only observed nodes and the value node remaining, either:

1. reduce some barren node;

2. reduce some probabilistic node which is unobserved; or

3. there is some decision node whose optimal policy can be selected and do so.

For example, consider the classic oil wildcatter problem shown in part a) of Figure 7 [Raiffa, 1968]. All of the nodes, Experimental Test, Seismic Test, Amount of Oil, Revenues, and Cost are unobserved and can be reduced in any order, to yield the diagram shown in part b). After optimal policy selection we obtain the diagram in part c). Now Test Results and Drill? are unobserved and can be reduced in any order to yield the diagram in part d). Finally, we can select the optimal policy for Test? and then reduce it to obtain the final diagram in part e).

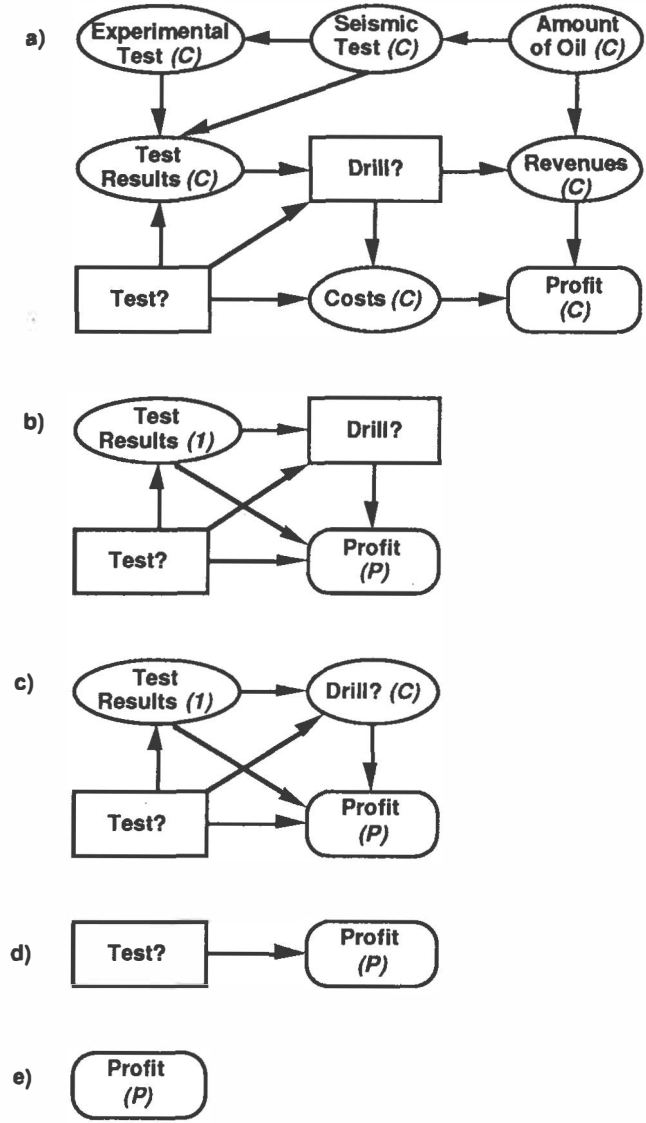

Figure 7. Oil Wildcatter Example of Decision Making in a Potential Influence Diagram

## 5 CONCLUSIONS AND FUTURE RESEARCH

We have shown that all of the properties of the potential influence diagram (PID) can be derived from two simple primitive operations, which we introduce in this paper. These operations help explain the different semantics of the PID as it relates to the conditional influence diagram (CID). The real power of the PID lies in the ease with which we can incorporate its efficiencies starting with an assessed CID, and still recover a CID for explanation and insight at any time.

Several research extensions involve incorporating some of the efficient techniques used to solve CID's into PID's, and they seem promising. First, it would be nice to

390   Shachter and Ndilikikesha

efficiently represent deterministic relationships in the PID without having to first compile the model [Andersen et al., 1989]. Second, because of its generality, the PID loses conditional independence information relative to the CID. One approach would be to maintain the CID independence information within the PID; another would be to maintain it separately. Finally, it seems straightforward to extend the CID dynamic programming algorithm [Tatman, 1985; Tatman and Shachter, 1990] to perform dynamic programming within the PID. This would combine the efficiency of the undirected computations [Shachter and Peot, 1992; Shenoy, 1991; Shenoy, 1992] with the flexibility and power of the directed graph. In these and in several other areas involving the tradeoffs between model transparency and computational efficiency, our real insights will come from practical experience in implementation and application.

## Acknowledgments

We appreciate the helpful comments of Bill Poland and the anonymous referees.